\documentclass{article}
\usepackage{arxiv}
\usepackage[utf8]{inputenc} 
\usepackage[T1]{fontenc}    
\usepackage[colorlinks=true,linkcolor=blue,urlcolor=blue,citecolor=black]{hyperref}       
\usepackage{url}            
\usepackage{booktabs}       
\usepackage{amsfonts}       
\usepackage{nicefrac}       
\usepackage{microtype}      
\usepackage{lipsum}		
\usepackage{graphicx}
\usepackage[backend=biber,style=apa,maxcitenames=2,sortcites=ynt]{biblatex}
\usepackage{doi}
\usepackage{xcolor}
\usepackage{comment}
\usepackage{lineno}
\usepackage{svg}
\usepackage{subcaption}
\usepackage{multirow}
\usepackage{authblk}
\usepackage{orcidlink}
\usepackage{rotating}

\makeatletter
\newcommand{\equalcontrib}[1]{\@fnsymbol{#1}}
\makeatother

\newcommand\blfootnote[1]{%
  \begingroup
  \renewcommand\thefootnote{}\footnote{#1}%
  \addtocounter{footnote}{-1}%
  \endgroup
}

\bibliography{OTEC-NLP.bib}

\title{Automating the analysis of public saliency and attitudes towards biodiversity from digital media}

\author[1]{\orcidlink{0000-0003-0292-5472}\,Noah~Giebink\textsuperscript{\equalcontrib{1}}}
\author[1]{\orcidlink{0000-0003-2643-5865}\,Amrita~Gupta\textsuperscript{\equalcontrib{1}}}
\author[3]{\orcidlink{0000-0001-6683-2078}\,Diogo~Ver\'{i}ssimo}
\author[2]{\orcidlink{0000-0002-1274-7721}\,Charlotte~H.~Chang}
\author[1]{\orcidlink{0000-0001-6683-20781}\,Tony~Chang}
\author[1]{\orcidlink{0000-0003-4360-0738}\,Angela~Brennan}
\author[1]{\orcidlink{0000-0003-4160-839X}\,Brett~Dickson}
\author[3]{Alex Bowmer}
\author[3]{Jonathan Baillie}

\affil[1]{\footnotesize Analytics Lab, Conservation Science Partners}
\affil[2]{\footnotesize Department of Biology, Environmental Analysis Program, Pomona College}
\affil[3]{On The Edge}

\renewcommand{\shorttitle}{Broadening conservation culturomics with NLP}

\hypersetup{
pdftitle={Public salience of biodiversity},
pdfsubject={conservation science},
pdfauthor={On The Edge, Conservation Science Partners},
pdfkeywords={conservation social science, environmental social media, NLP},
}

\begin{document}
\maketitle

\blfootnote{\textsuperscript{\equalcontrib{1}} Equal contribution.}
\blfootnote{Correspondence to: Amrita~Gupta~\texttt{<agupta375@gatech.edu>},
\\\hspace*{90pt}Diogo~Ver\'{i}ssimo~\texttt{<diogo.gasparverissimo@biology.ox.ac.uk>}
\\\hspace*{90pt}Charlotte~Chang~\texttt{<chchang@pomona.edu>}, 
}

\begin{abstract}
 Measuring public attitudes toward wildlife provides crucial insights into our relationship with nature and helps monitor progress toward Global Biodiversity Framework targets. Yet conducting such assessments at a global scale presents challenges. Manual curation of search terms for querying mass media (news) and social media is tedious and costly, and can lead to potentially biased results. Raw news and social media data returned from queries are often cluttered with irrelevant content and syndicated, or republished, articles. We aim to overcome these challenges associated with monitoring public engagement with biodiversity at scale by leveraging modern Natural Language Processing (NLP) tools. We introduce a folk taxonomy approach for less biased and more efficient search term generation. Additionally, we employ cosine similarity on Term Frequency-Inverse Document Frequency vectors to identify and filter syndicated articles. We introduce an extensible relevance filtering pipeline which uses unsupervised learning to reveal common topics, followed by an open-source zero-shot Large Language Model (LLM) to assign topics to news article titles, which are then used to assign relevance. Finally, we conduct sentiment, topic, and volume analyses on resulting data. To illustrate our methodology, we conduct a case study of news and X (formerly Twitter) data before and during the COVID-19 pandemic for various mammal taxa, including bats, pangolins, elephants, and gorillas. During the data collection period, up to 62\% of articles mentioning bats were deemed irrelevant to biodiversity, underscoring the importance of relevance filtering. At the pandemic's onset, we observed increased volume and a significant sentiment shift toward horseshoe bats, which were implicated in the pandemic, but not for other focal taxa. The proposed methods open the door to conservation practitioners applying modern and emerging NLP tools, including LLMs “out of the box,” to analyze public perceptions of biodiversity during current events or campaigns.
\end{abstract}

\keywords{Conservation social science \and Environmental social media \and Natural language processing}

\linespread{1.5}

\section{Introduction}

Public interest in biodiversity is pivotal to the success of conservation efforts, but varies significantly across species, geographies, and time. While targeted conservation campaigns can amplify public engagement around focal species and catalyze policy change~\parencite{thaler2017lions}, the systemic change needed to halt biodiversity loss requires cultivating public awareness and support for nature and biodiversity as a whole~\parencite{diaz2019,CBD2022KumingMontreal}.

Monitoring public attitudes towards species comprehensively and at scale is a formidable challenge, but conservation culturomics--analyzing digital data to examine societal relationships with nature--holds great promise for this purpose~\parencite{correiaDigitalDataSources2021,ladle2016}.

Digital data sources offer global reach and cost-efficiency over conventional opinion-based surveys, and can reveal information-seeking behavior rather than behavioral intent~\parencite{cooper2019}. Although recent work has developed attention metrics based on Wikipedia page views~\parencite{millard2021a,vardi2021} and Google Trends~\parencite{cooper2019,burivalova2018,vardi2021}, news and social media offer additional insights into the context of public attention on species~\parencite{roberge2014}. News media narratives shape public perceptions~\parencite{king2017news} while social media have become a dominant platform for sharing news and viewpoints toward issues including biodiversity conservation \parencite{changBioScience22,verissimo2021,papworth2015}. However, unlike Google Trends and Wikipedia page views, news and social media yield unstructured text data, requiring careful search and filtering for relevant content.

Selecting effective search terms for species in keyword-based search application programming interfaces (APIs) is a nuanced task. This partly stems from the mismatch between the specialized biological nomenclatures conservation experts use, such as Latin (e.g., \emph{Rhinolophus affinis}) or specific common names (e.g., ``Intermediate horseshoe bat''), and the broader folk taxonomic terms the public favors (e.g., ``bat'' or ``horseshoe bat'') that may encompass multiple related species~\parencite{beaudreau2011}. This highlights a trade-off between specificity and volume of relevant content when assessing public views on species groups. Using Latin~\parencite{jaric2020,ladle2019culturomics} or full common names~\parencite{roberge2014,kulkarniAutomatedRetrievalInformation2021} as keywords enhances specificity but risks overlooking general references to species within folk taxonomies, potentially biasing search results towards scientific content, especially for species lacking well-known common names. Conversely, common names for folk taxa are challenging to infer. Past efforts hand-curated common names for target taxa~\parencite{fink2020}, but extending this approach to thousands of species is both arduous and subjective. Additionally, some common names (e.g., ``elephant'') appear as substrings within unrelated species names (e.g., ``elephant seal''), requiring careful consideration when constructing search queries.

Another challenge in conservation culturomics is the use of species common names in non-biological contexts, such as sports teams (e.g., Clemson Tigers), individuals (e.g., Tiger Woods), and other entities. Machine learning and natural language processing (NLP) approaches can be used to develop text classification models for filtering out such irrelevant results. These models predict whether or not a sample of text pertains to biodiversity conservation~\parencite{kulkarniAutomatedRetrievalInformation2021}, target species or conservation topics~\parencite{keh2023NewsPanda,hunter2023using,roll2018using,egriDetectingHotspotsHumanWildlife2022}. However, they require extensive manually annotated data for training, are susceptible to biases in data labeling, and may not generalize well to examples not seen during training.

To address these challenges, we develop a pipeline for retrieving online news and X (formerly Twitter) posts about biological taxa of conservation interest. We introduce a novel method for deriving a folk taxonomy from English common names via substring matching, simplifying the identification of names used in everyday language to refer to animals. This approach facilitates analysis of less well-known species by grouping them into more broadly recognized taxa, overcoming the limitations posed by using only Latin or full common names for these species. It also reveals spurious groupings of unrelated species, corrected by incorporating negative search terms into API queries to enhance search specificity. Furthermore, we use a zero-shot text classification model to filter out irrelevant content, a cutting-edge machine learning approach that obviates the need for data annotation by generalizing to new tasks without additional training. We illustrate the utility of our pipeline in an example analysis of public discourse on several mammal taxa from 2019 to 2021, encompassing periods both before and after the United Nations World Health Organization officially declared the COVID-19 pandemic on March 11, 2020. Early in the outbreak, interest in wildlife increased, particularly in potential zoonotic coronavirus sources like bats or pangolins~\parencite{vijay2021,petrovan2021,zhou2020pneumonia}. We explore changes in public perceptions toward bats and pangolins (versus elephants and gorillas, which were not implicated in the pandemic) by examining discourse volume and sentiment shifts over time.

\section{Materials and Methods}
\label{sec:methods}

\begin{figure}
	\centering
	\includegraphics[width=\textwidth]{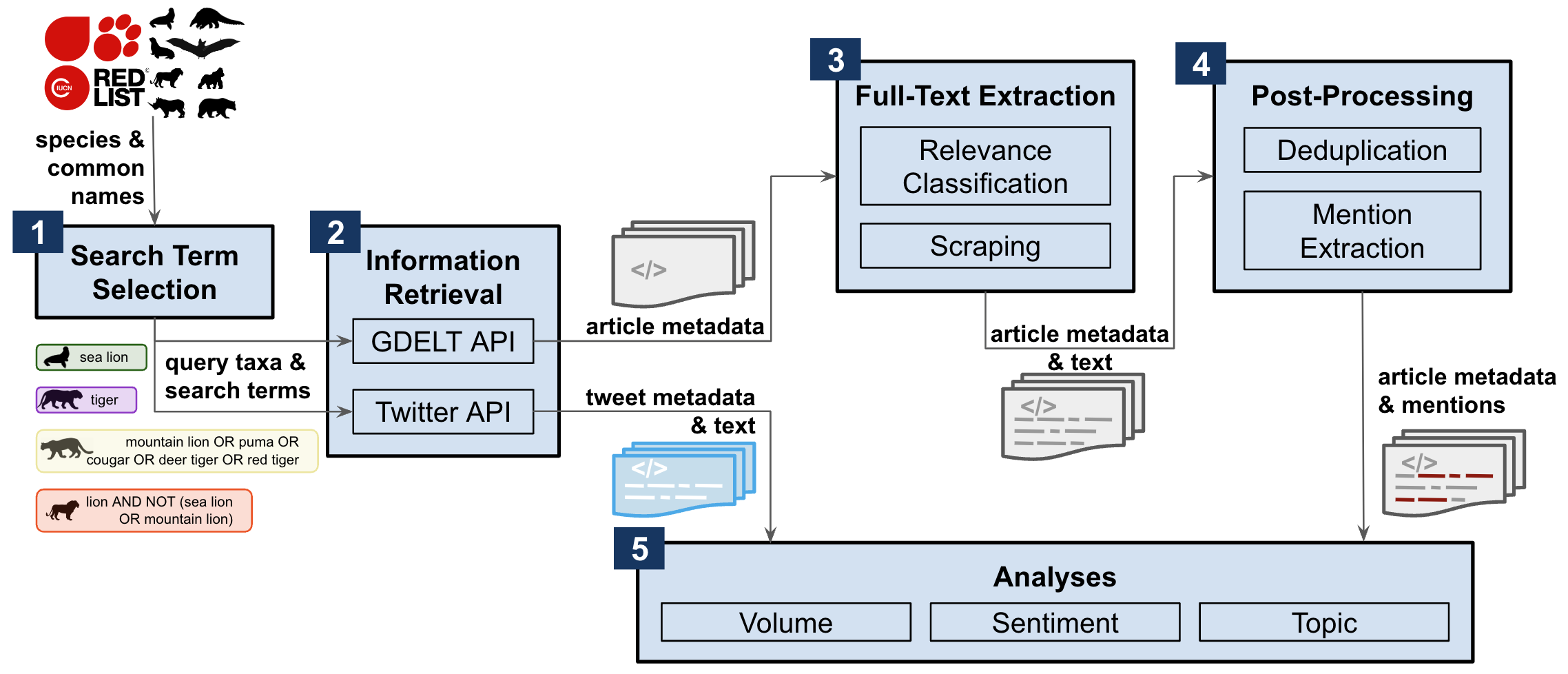}
	\caption{A diagram of the data pipeline, starting from constructing a folk taxonomy to derive search terms; retrieving news and tweets by querying each data source; performing zero-shot relevance modeling and scraping to obtain full-text for the news media articles; filtering out syndicated news and identifying specific references to queried taxa within news articles; and finally conducting analyses on shifts in volume, sentiment, and topics in the tweets and news articles through time and over space.}
	\label{fig:data-pipeline}
\end{figure}

Our pipeline for collecting online news articles and social media posts about biological taxa of interest is illustrated in Figure~\ref{fig:data-pipeline} and summarized below:
\begin{enumerate}
    \item \textbf{Query taxa and search term selection}: We begin by selecting the taxa for analysis, focusing on either individual species or broader categories based on public visibility. This sets the foundation for our data collection by specifying which species are encompassed in each targeted search. Details can be found in Section~\ref{subsec:methods/search-term-selection}.
    \item \textbf{Information retrieval}: Following the identification of target taxa and corresponding search terms in the previous step, we use keyword search APIs to retrieve online news articles and social media posts containing search terms related to each query taxon (refer Section~\ref{subsec:methods/information-retrieval}).
    \item \textbf{Full text extraction}: For news articles, where our initial retrieval yields only titles and URLs, we first classify these article titles by topic to determine their relevance to conservation (see Section~\ref{subsubsec:methods/relevance-filtering}). Only articles deemed relevant undergo full text scraping (Section~\ref{subsubsec:methods/scraping}), ensuring efficiency by avoiding the extraction of text from irrelevant articles.
    \item \textbf{Data post-processing}: We apply text similarity techniques to identify and filter out syndicated articles, which are near-duplicates of original content and could introduce redundancy into our text corpus. Further, we extract specifically those sections of text with original articles that directly reference the target taxa, thus enhancing the specificity of our analysis.
    \item \textbf{Data and text analysis}: We leverage the collected data for a range of analyses aimed at uncovering insights into the public discourse surrounding the target taxa. We explore the volume of online content about different target taxa and how that varies geographically and over time. Sentiment analysis can help track shifts in the tone of these discussions, while topic analysis sheds light on underlying themes in these discussions. These examples illustrate the versatility of our dataset in facilitating diverse analytical approaches to deepen our understanding of the discourse dynamics related to the target taxa.
\end{enumerate}

\subsection{Search term selection}
\label{subsec:methods/search-term-selection}

\begin{figure}
	\centering
	\includegraphics[width=\linewidth]{"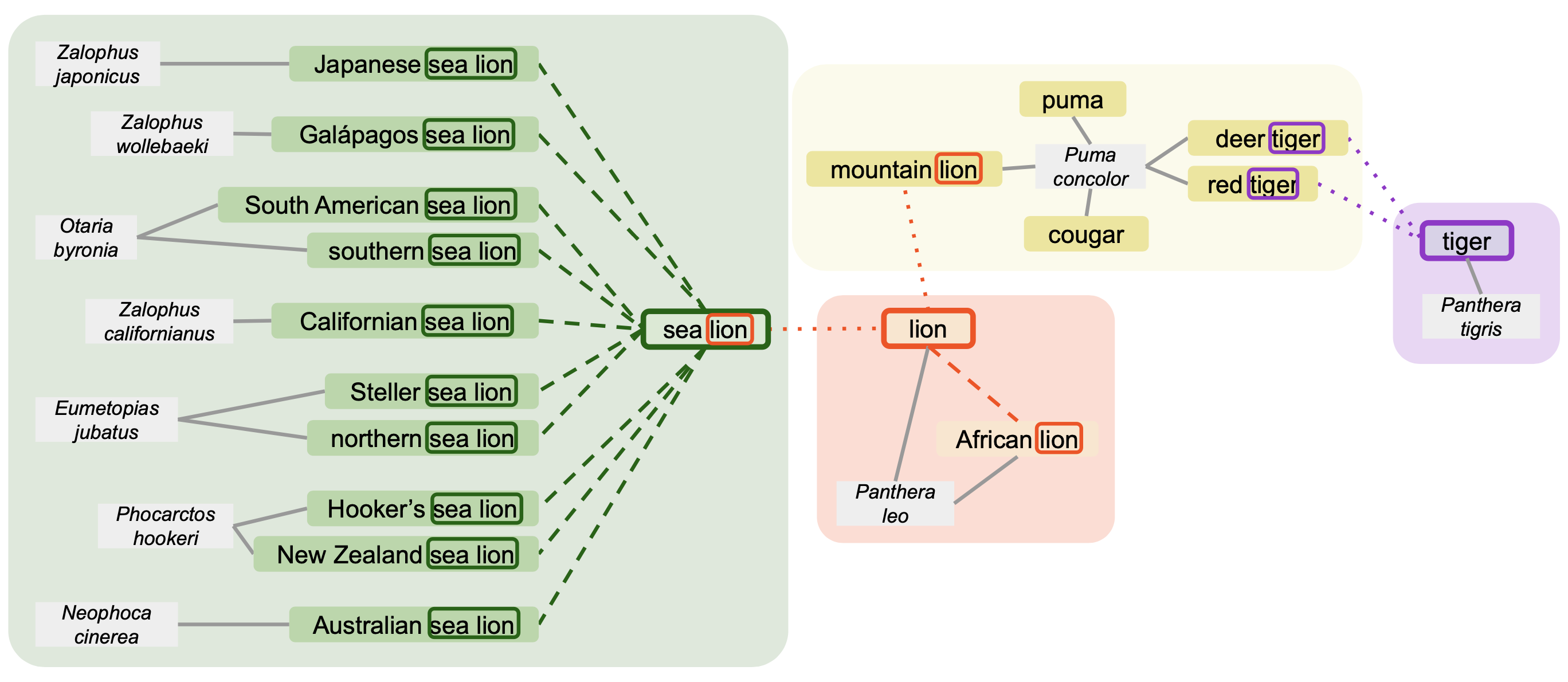"}
	\caption{Example of an initial connected component in the folk taxonomy graph for species in Order Carnivora based on their IUCN Red List common names. Solid lines represent edges between species and their listed common names; dashed lines represent edges between names and simplified names; and dotted lines represent connections that would be pruned on inspection to separate conceptually distinct taxa.}
	\label{fig:folktax}
\end{figure}

Identifying salient folk taxa--groups of species as referenced in everyday language--is a fundamental step in monitoring public perceptions of these taxa in conservation contexts. We accomplished this through a human-in-the-loop approach, using English-language common names for species and their simplified forms as the basis for identifying these taxa in our analysis. First, we gathered the comprehensive list of mammalian species and their English common names from the International Union for Conservation of Nature and Natural Resources (``IUCN'') Red List (\href{https://www.iucnredlist.org/}{IUCNredlist.org}), encompassing a total of 5,650 species and 9,150 common names. We leveraged an efficient dynamic programming algorithm to extract shared trailing substrings from the common names (such as "sea lion" from "South American sea lion" and "Californian sea lion", see Fig.~\ref{fig:folktax}), yielding prospective folk taxa. We then constructed an undirected graph representation of the connections between species, their common names, and the identified shared substrings. We clustered this graph into connected components, each of which represents a candidate taxon comprised of a group of species and a simplified set of names for them. Each cluster was manually inspected to ensure that the species formed a coherent group. Otherwise, nodes or edges in the graph were modified before repeating the clustering and inspection.

In some cases, our method grouped several taxa into a broader taxon that might be considered too coarse. For instance, the ``Andean bear'', ``black bear'', ``brown bear'', ``polar bear'', ``sloth bear'', and ``sun bear'' were initially grouped under the ``bear'' taxon. Given the widespread recognition of distinct bear species, one could consider eliminating the node associated with the shared substring ``bear'' to separate these species into distinct clusters. A more complex issue arises, however, when shared substrings are found between common names of unrelated species.
For instance, the substring ``lion'' appears in the common names for \emph{Panthera leo} (``lion''), \emph{Puma concolor} (``mountain lion''), \emph{Leontopithecus spp.} (``lion tamarins''), \emph{Zalophus wollebaeki} (``Galápagos sea lion''), and other unrelated species (Fig.~\ref{fig:folktax}). Conducting a search using the term ``lion'' could potentially yield results encompassing all these taxa. To avoid this, we incorporated negative keywords (e.g. ``lion'' AND NOT ``mountain lion'' AND NOT ``sea lion'' AND NOT ``lion tamarin'') to improve differentiation among these species during searches, a strategy not documented in prior work.

For each folk taxonomic entity, we compiled a set of positive keywords, at least one of which must be present in a search result, and an optional set of negative keywords, all of which must be absent from a search result. Additional details about the graph construction can be found in Supplementary Information Section \ref{subsec:appendix-folktax-graph}.

\subsection{News and social media information retrieval}
\label{subsec:methods/information-retrieval}

We collected online news articles from the Global Database of Events, Language, and Tone (GDELT), a live database capturing global news media offering full-text search via the GDELT 2.0 DOC API. Using positive and, where applicable, negative search keywords for each target taxon, we requested English-language articles published between January 1, 2019 and December 31, 2021. Each query returned JSON-formatted article metadata that included the article's title, URL, domain, date, and country of publication. To work within the limit of 250 results per query imposed by the API, we divided the three year period into shorter intervals, aggregating results from each interval to form our final dataset.

Similarly, for social media analysis, we utilized the Twitter Academic Access v2 API to access Twitter's full archive of public tweets. We queried this API with the positive and negative keywords for each target folk taxonomic entity, requesting only tweets written in English and including geolocation data to support analyses on geographic differences in species media portrayals. Twitter data collection concluded before February 9, 2023, ahead of potential deprecation notices for the Academic Access API by Twitter.

\subsection{News full-text extraction}
\label{subsec:methods/full-text-extraction}

\subsubsection{Relevance filtering}
\label{subsubsec:methods/relevance-filtering}
The keyword-based search described above often retrieves a mixture of relevant and irrelevant results for wildlife conservation~\parencite{kulkarniAutomatedRetrievalInformation2021}. For instance, a query using the search term ``tiger'' might fetch articles mentioning sports teams (e.g. the Clemson Tigers), people (e.g. Tiger Woods or Tiger Shroff), companies (e.g. Tiger Global Management, LLC), places (e.g. Tiger Hill), or even events (e.g. Year of the Tiger). Articles in which the search keywords refer to non-animal entities should be excluded from the corpus of wildlife-focused news articles. However, GDELT queries return only metadata such as titles and URLs, not full texts, requiring us to decide whether an article likely uses the search keywords in the intended sense from these relatively limited metadata.

We make the simplifying assumption that articles about topics related to nature and conservation are more likely to use keywords in the intended context. Our goal, then, is to classify the title of a news article as relevant or irrelevant to wildlife or conservation. We developed a topic classification-based approach, in which an online news article is predicted as belonging to one or more predefined topics, a subset of which are considered relevant. We derived the set of predefined article topics in a two-stage approach. In the first stage, we randomly sampled 10,000 articles from GDELT query results for news articles from 2019, stratified such that at least one taxon from each of 14 Mammalian Orders was represented. The resulting sample contained mentions of 154 mammalian taxa. The full-text of these articles was obtained via webscraping (see Section~\ref{subsubsec:methods/scraping}) and text snippets containing animal search terms were extracted, with each snippet being 7 sentences long 
(for context, 3 sentences before and after the sentence mentioning the taxon). We used Latent Dirichlet Allocation (LDA) to perform unsupervised topic modeling on these text snippets, obtaining 40 initial topics. LDA models texts as mixtures of topics, which are themselves mixtures of words, allowing for the discovery of underlying thematic structures in large text corpora. The choice of 40 topics was numerous enough to glean many informative topics without exceeding the model's capacity to reliably converge within 150 iterations. We reviewed the 20 words scored most important by the model for each topic to assign a semantically meaningful label to each one, yielding 23 topic labels which we then grouped into relevant versus irrelevant topics as follows:
\begin{description}
    \item \textbf{Relevant:} agriculture, climate change, conservation, energy, health, infrastructure, natural disasters, nature, outdoor recreation, science and technology, tourism, wildlife, habitat loss, invasive species, pollution
    \item \textbf{Irrelevant:} business, crime, education, entertainment, food, holidays, politics, sports
\end{description}

We defined relevant topics as those that discussed species in a biological, conservation, or real-world context, whereas irrelevant topics were instead focused on non-biological issues.

Given these predefined topics, our next challenge was to classify GDELT query results among these predefined topics, keeping in mind that we have access to only the article title at this stage in the GDELT data collection pipeline. We used Facebook's Bidirectional and Autoregressive Transformers (``BART'') model to perform multi-label ``zero-shot'' topic classification for the GDELT article data using the topics identified through our LDA analysis of the full-text subset dataset \parencite{lewis2019}. Each article title received a unit-sum vector of topics with probabilities across the 23 topics enumerated above. If an article title was predicted as having any of the relevant topics with a model score greater than 0.5, the article was considered relevant and was flagged for webscraping. The zero-shot BART model is capable of predicting topics on new data, given the extensive scale of the data that were used to train these models and their watershed advance in creating numeric representations (also known as ``embeddings'') that can capture the semantic structure of the English language. The major advantage of these models is that they enable conservation practitioners to now filter text corpora that would simply be impossible to manually review.

\subsubsection{Scraping the full text of articles}
\label{subsubsec:methods/scraping}
To obtain the full text of news articles flagged as relevant, we first submitted an HTTP request for the HTML content of each relevant GDELT news article URL. If the request was successful, the HTML content was parsed using one of three Python libraries (\texttt{trafilatura}, \texttt{newsplease}, or \texttt{boilerpy3}) to extract the article body. Often, however, the HTML request or the text extraction was unsuccessful due to broken URLs. As a method of recourse in these cases, we searched for a snapshot of the article on the Internet Archive. If a snapshot was found, we requested the HTML content of this snapshot and attempted to extract the article body text using the same combination of Python libraries as before.

\subsection{News data post-processing}
In mainstream media, news articles are often syndicated across multiple outlets with minimal changes to the text~\parencite{kulkarniAutomatedRetrievalInformation2021}. To prevent bias in downstream models and avoid redundant analyses on near-identical content, we implemented a process to identify duplicates. We measured the similarity between articles by first using Term Frequency-Inverse Document Frequency to create a vector representation of each article's text based on its most distinctive words, and then computing the cosine similarity between pairs of article vectors. We compared all pairs of articles published within two months of each other, as syndicated articles are typically released soon after their originals. If the cosine similarity exceeded 0.95, indicating a high degree of similarity, we classified the later-published article as a syndicate of the earlier one. Conversely, if an article's cosine similarity score with every other article published within the preceding two months was below 0.95, we classified it as an original.

Next, we isolated sentences within articles that directly reference the target taxa, a step we call ``entity mention detection''. This step enables us to precisely apply NLP tasks like sentiment analysis and topic modeling to text segments containing the entity of interest. This is especially useful for longer bodies of text like articles, which can discuss many different things and have shifts in tone. We scanned each article for the positive search terms for a target taxon. Upon finding a mention, we extract the sentence that contains this reference along with the sentence immediately preceding it. Including the preceding sentence is helpful as it often frames the mention with additional context.

\subsection{Analyzing public discourse about species}
For each taxon, we had a set of articles and tweets from 2019 through the end of 2021. To determine the overall volume of discourse toward each taxon, we aggregated the number of articles mentioning each taxon by month and country. 

We examined the sentiment of media and public discussion of species using a lexicon sentiment model. Specifically, we used the ``Valence Aware Dictionary and sEntiment Reasoner'' (abbreviated to VADER, \parencite{hutto2014vader}).
This yielded a sentiment score for each article ranging from -1 (negative) to 0 (neutral) to 1 (positive). We also aggregated these article-level sentiment scores by month and country to examine patterns in public discourse regarding species.

We illustrate how conservation social science researchers and practitioners interested in messaging or marketing to conserve biodiversity can perform different analyses using the outputs of our data pipeline. Using information on the country where each news article is published, we show how one can create choropleth maps of the volume of public discourse toward different taxa. We use chord diagrams to visualize the distribution and co-occurrence of different topics associated with news media coverage of each taxa. Finally, we use breakpoint analyses \parencite{killick2014} to evaluate whether or not there were significant changes in the mean volume or sentiment through time for different taxa.

\section{Results}

\subsection{Creating a folk taxonomy and collecting data on target taxa}
\begin{figure}[th]
	\centering
	\includegraphics[width=\textwidth]{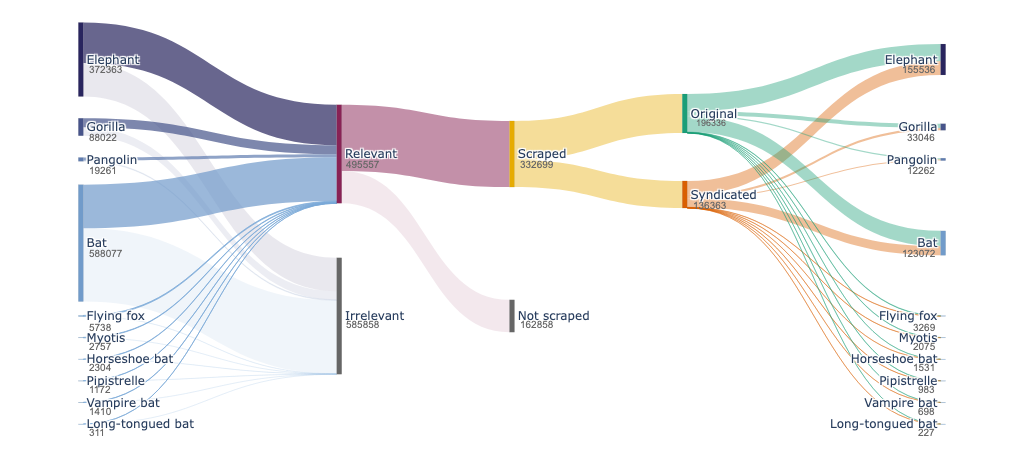}
	\caption{Number of news articles obtained at each stages in the GDELT data collection pipeline run fully on ten query taxa, from querying, to relevance filtering, webscraping, and deduplication.}
	\label{fig:gdelt-pipeline}
\end{figure}
Using the approach described in Section~\ref{subsec:methods/search-term-selection}, we derived folk taxonomic terms for the species within each of the 26 Orders of mammals in the IUCN Red List. Figure~\ref{fig:folktax} illustrates one of the connected components for the Order Carnivora before inspection, conveniently grouping several species under the folk taxon ``sea lion''--a term the public is more likely to use than any of those species' full common names. However, it also reveals links between `lion' and both `sea lion' and `mountain lion' due to the shared substring `lion.' These links, which could lead to mixed search results when searching for 'lion,' highlighted the need for negative search terms to improve search specificity. We conducted comprehensive analysis using our proposed pipeline for 10 taxa, which are listed along with their corresponding scientific taxa in Table~\ref{tab:folktax}. These range from the Genus level (gorilla) to the Order level (bats). We also considered more specific yet popularly recognized taxa like ``flying fox'' and ``vampire bat'' and lesser-known but still distinct taxa like ``pipistrelle'' and ``horseshoe bat''.

Figure~\ref{fig:gdelt-pipeline} shows the outcome at each stage of our GDELT data collection pipeline applied to the 10 case study taxa. Raw article counts varied from 588,077 results for ``bat'' to 311 for ``long-tongued bat''. Notably, approximately 54\% of articles were predicted to be irrelevant to wildlife, with ``bat'' (62.6\%), ``gorilla'' (48.4\%), ``elephant'' (45.3\%) and ``vampire bat'' (36.4\%) yielding high proportions of unrelated content due to homonymy (e.g. ``bat'' as a piece of sports equipment), idiomatic expressions (e.g. ``elephant in the room'', ``800 pound gorilla'', and ``off the bat''), and popular culture depictions of these animals. Full-text scraping was attempted for all articles that were predicted to be relevant based on their title, yet about a third were inaccessible due to broken links. We also found that 41\% of articles across these taxa were syndicated, indicating significant potential for computational efficiency by limiting analyses to only relevant, original articles (Table~\ref{tab:relevance}). Ultimately, taxa with widespread popular appeal (elephants, gorillas) had more wildlife news articles than lesser-known taxa (pangolins), and generic taxa had more articles than specific ones.

Pivoting to social media, we observed that the public made anywhere between several hundred to nearly 300,000 posts about different taxa from 2019 to 2022 (Table~\ref{tab:twittercts}). In our subsequent analyses, we now illustrate different use cases of the data generated by our data pipeline.

\begin{table}
        \centering
	\caption{The counts across taxa for Twitter posts from 2019-2021.}
	\label{tab:twittercts}
	\begin{tabular}{lr}
		\toprule
		Entity & Count \\
		\midrule
		Elephant & 171059 \\
		Gorilla & 52650 \\
		Pangolin & 3667 \\
		Bat & 325319 \\
		Flying fox & 1889 \\
		Myotis & 166 \\
		Horseshoe bat & 140 \\
		Pipistrelle & 354 \\
		Vampire bat & 462 \\
		Long-tongued bat & 80 \\
		\bottomrule
	\end{tabular}
\end{table}

\subsection{How does discourse vary around the world?}

\begin{sidewaysfigure}
    \centering
    \begin{subfigure}{0.98\textwidth}
        \includegraphics[width=\textwidth]{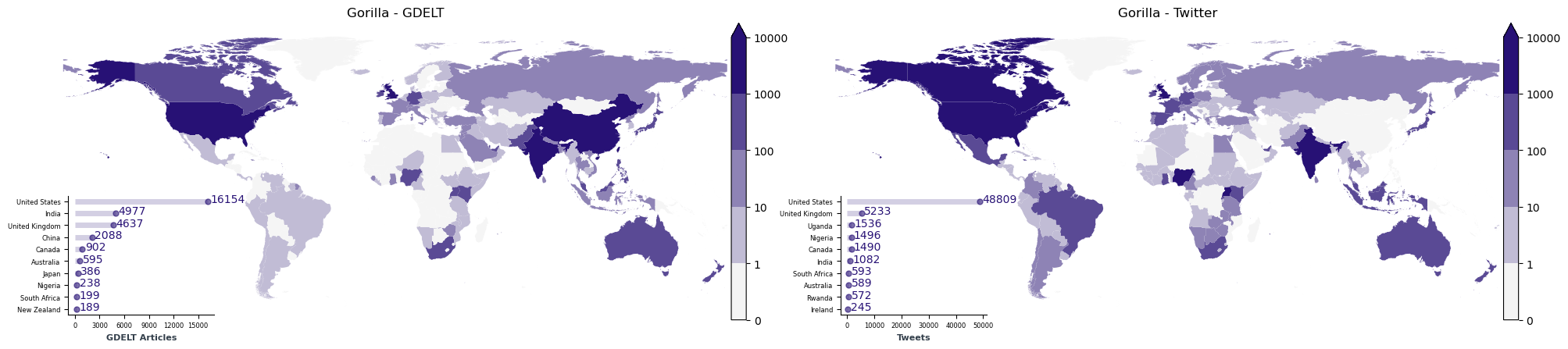}
        \caption{Gorilla}
        \label{fig:gorilla-choropleth}
    \end{subfigure}
    \begin{subfigure}{0.98\textwidth}
        \includegraphics[width=\textwidth]{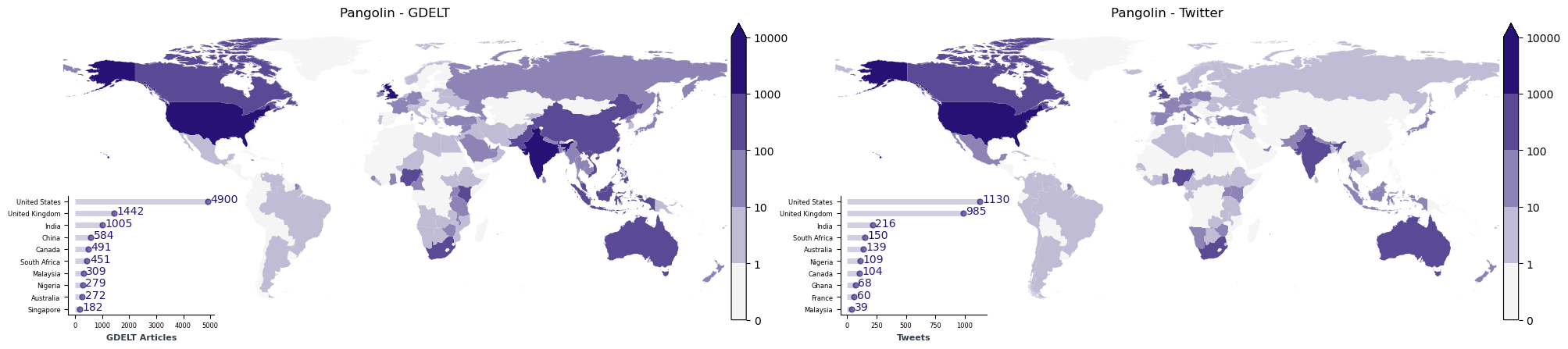}
        \caption{Pangolin}
        \label{fig:pangolin-choropleth}
    \end{subfigure}
    \begin{subfigure}{0.98\textwidth}
        \includegraphics[width=\textwidth]{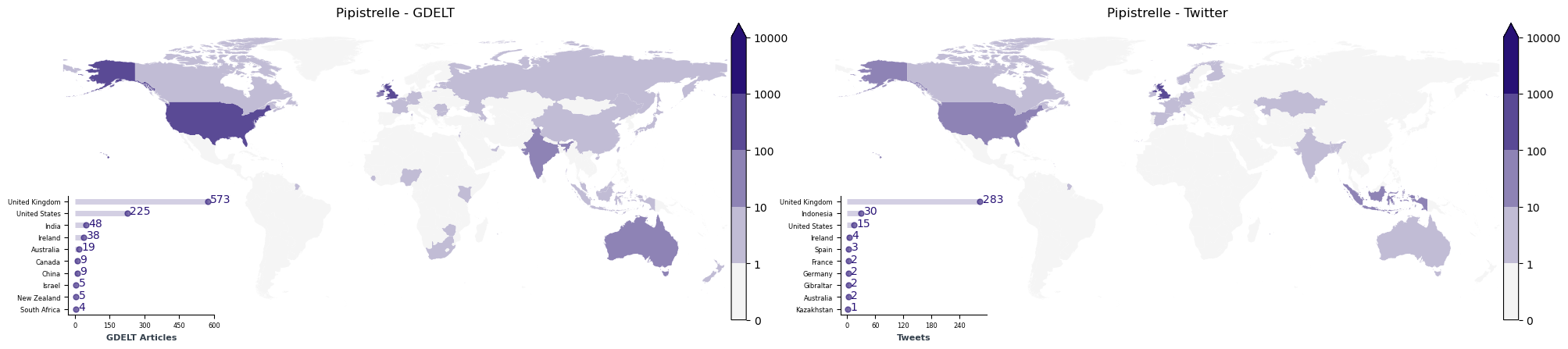}
        \caption{Pipistrelle}
        \label{fig:pipistrelle-choropleth}
    \end{subfigure}

\caption{Volume of relevant news articles from GDELT (left) and Tweets (right) between January 1, 2019 and December 31, 2022 for gorillas (\ref{fig:gorilla-choropleth}), pangolins (\ref{fig:pangolin-choropleth}), and pipistrelles (\ref{fig:pipistrelle-choropleth}). Insets show top 10 countries by volume.}
\label{fig:volume-choropleths}
\end{sidewaysfigure}

Figure~\ref{fig:volume-choropleths} shows spatial variations in media coverage across different taxa. Globally recognized animals like gorillas receive widespread attention online, while less well-known taxa such as pangolins and pipistrelle bats see more geographically concentrated coverage. Pangolins are primarily featured in Southeast Asia, whereas pipistrelle bats, despite their prevalence in the British Isles and widespread distribution in Asia, attract less media attention outside the UK. These findings show media exposure to different animal taxa varies by geography, potentially influencing levels of awareness and familiarity.

\subsection{Which topics are associated with different taxa?}

\begin{figure}
	\centering
	\includegraphics[width=\textwidth]{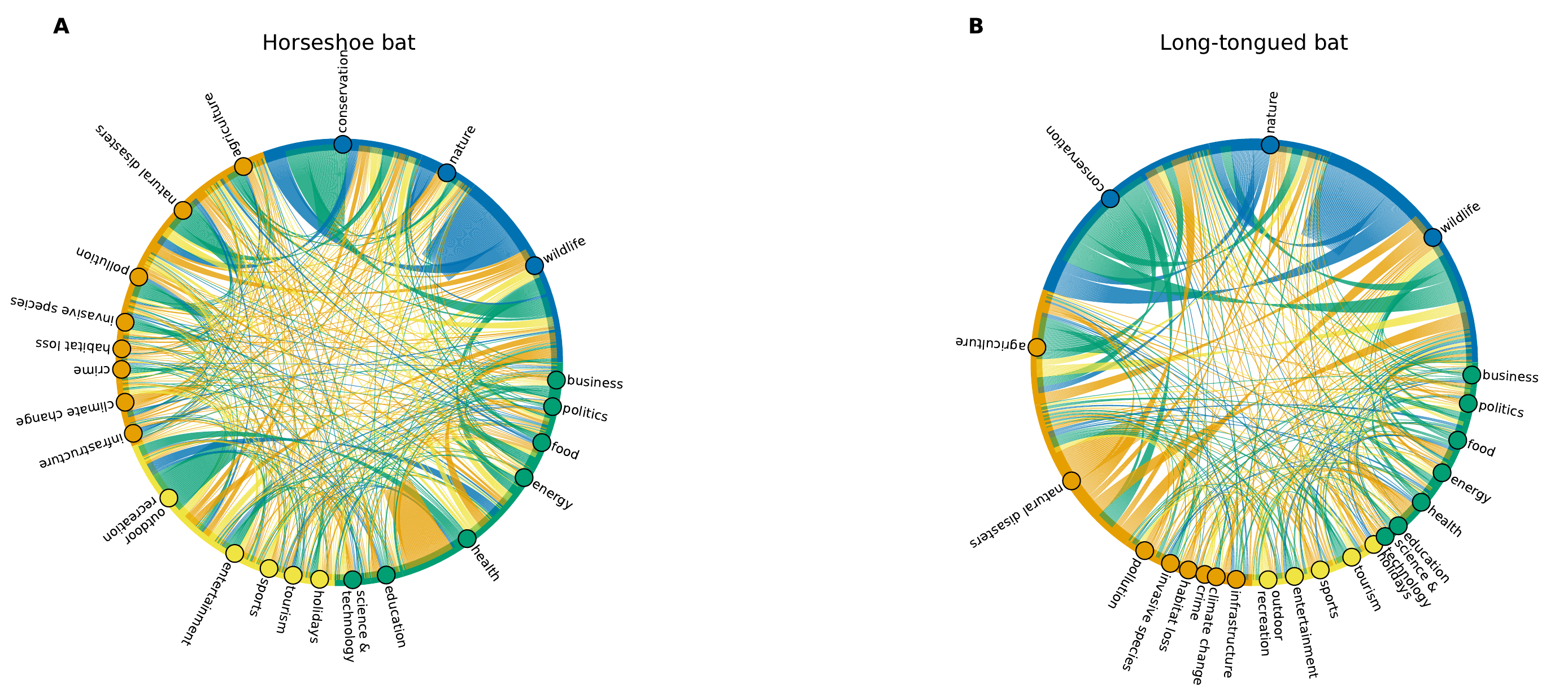}
	\caption{Chord diagrams depicting the co-occurrence of relevant topics for two focal taxa, Horseshoe bat (family Rhinolophidae) and Long-tongued bat (genera Glossophaga, Craseonycteris, and Leptonycteris). A wider chord indicates that more articles contain two taxa and each chord is a band colored by one of the nodes that is being connected. The circular perimeter of the chord displays the proportional occurrence of each topic in the dataset, and the colors correspond to different groups of topics.}
	\label{fig:chordBat}
\end{figure}

The advent of large language models trained on Internet-scale text corpora offers major advances for zero-shot learning, where practitioners can use the predictions of a model on their own datasets to help sift through volumes of data that simply defy manual review. Figure~\ref{fig:chordBat} displays the co-occurrence of topics predicted by Facebook's BART (Bidirectional and Auto-Regressive Transformers) model \parencite{lewis2019}. Each line in one of the chord diagrams represents a topic that is co-occurring in an article with another topic (for the full set of chord diagrams for all folk taxonomic entities, please refer to Figure~\ref{fig:relevantTopics}).

Practitioners could use approaches such as these to evaluate how wildlife is framed in the news media. Comparing horseshoe bats, a known reservoir of SARS-CoV, versus long-tongued bats, which are not regarded as a coronavirus reservoir, we observed that the distribution and co-occurrence of topics was quite different between these two groups of species. Long-tongued bats had much more news coverage devoted to nature-based topics such as conservation or wildlife. Moreover, the nature or conservation threats topics (e.g. nature, wildlife, climate change, habitat loss, etc.) tend to occur with one another in articles. In contrast, horseshoe bat media coverage exhibited comparatively more discourse on the topics of conservation threats (e.g. habitat loss, natural disasters, climate change) or socio-economic issues (e.g. business, health, education). Health and food were more prevalent topics for horseshoe bats compared to long-tongued bats. For both types of bat, however, the chord diagram indicates that there is substantial co-occurrence of different topics at the level of individual articles.

\subsection{How has the salience of taxa changed through time?}

\begin{figure}
	\centering
	\includegraphics[width=0.9\textwidth]{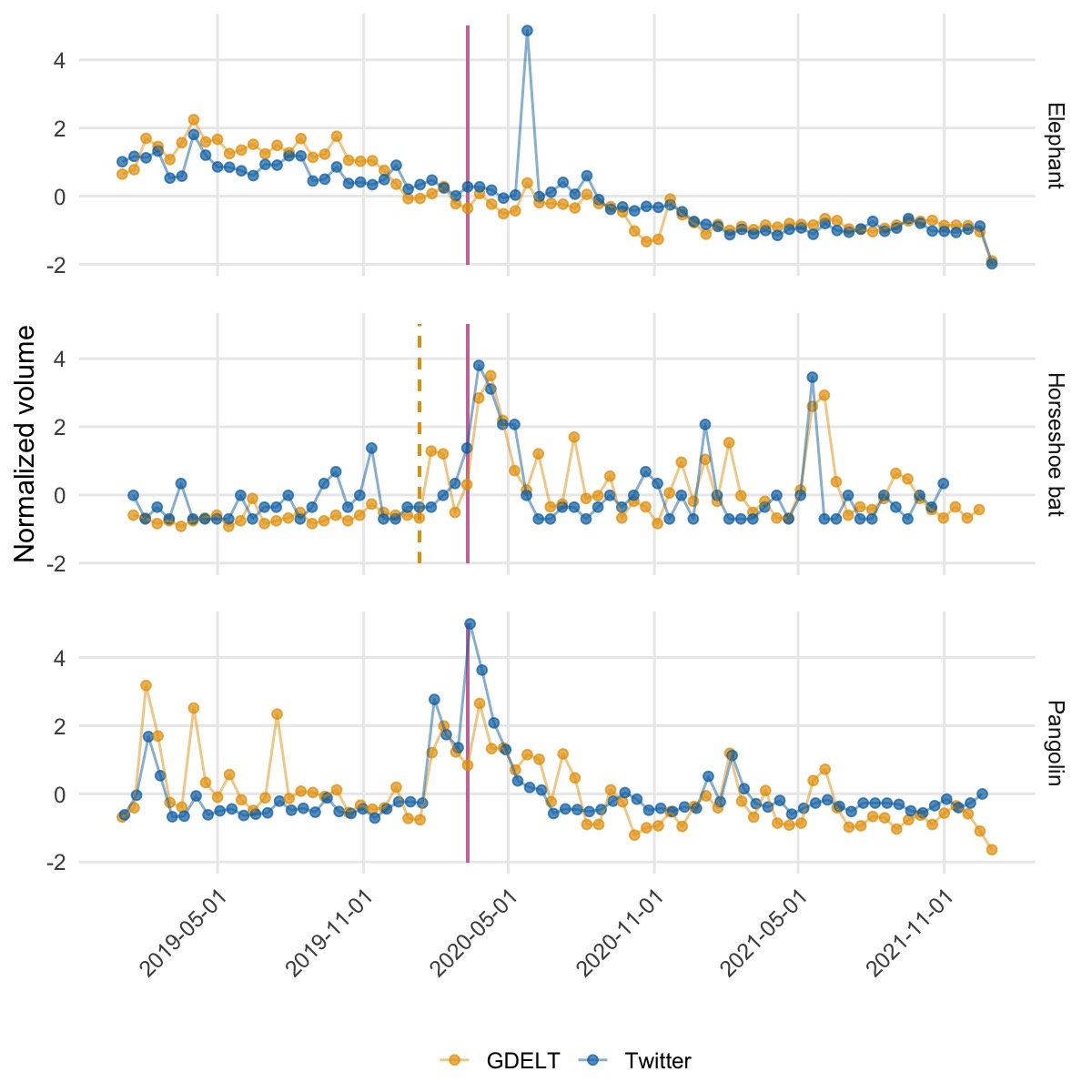}
	\caption{Changes in volume through time. The solid vertical magenta line denotes March 11, 2020, which was the date when the UN WHO declared COVID-19 a pandemic. The dashed vertical orange or blue lines correspond to any significant breakpoints in the trend for GDELT or Twitter respectively, after conducting Bonferroni family-wise error correction.}
	\label{fig:volumeTime}
\end{figure}

Figure~\ref{fig:volumeTime} displays changes in the volume of mass media articles referencing different taxa, shown using normalized counts for each taxa's count of articles or Twitter posts, aggregated over a two week sliding window. Comparing taxa implicated as coronavirus hosts or as potential spillover hosts (pangolin or horseshoe bat) versus species of conservation concern that are not clearly associated with COVID-19 (e.g. elephant), we observed differences in the salience of these taxa.

A breakpoint analysis indicated that there were significant changes in the volume of news media articles published on horseshoe bats. Specifically, there were an average of 3 articles published every two weeks on horseshoe bat before January 10, 2020, and this volume jumped to an average of 20 articles every two weeks after this breakpoint. We did not observe any other significant breakpoints for the count of Twitter posts or the count of news media articles for any of the other focal taxa. However, in the full set of taxa (Figure~\ref{fig:SIvolume}), we found breakpoints in news media coverage for flying fox, gorilla, and myotis between November 2019 to September 2020. We did not find any breakpoints for any taxa in terms of the volume of Twitter posts in our sample. However, unlike the horsehoe bat, all of these taxa exhibited either no change or a reduction in average volume in news media content. 

\subsection{How has the sentiment of discourse about taxa changed through time?}

\begin{figure}
	\centering
	\includegraphics[width=0.9\textwidth]{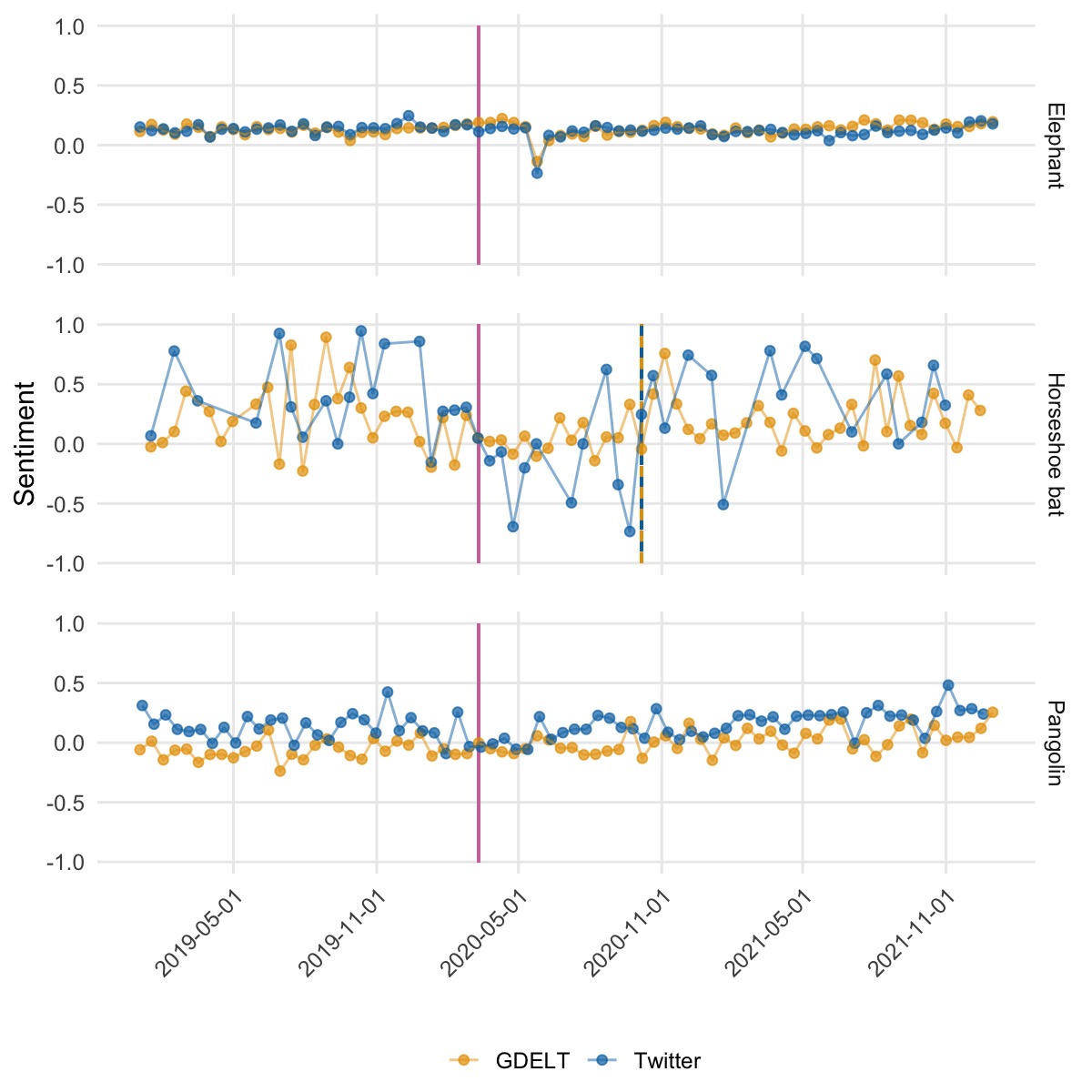}
	\caption{Changes in sentiment through time. The solid vertical magenta line denotes March 11, 2020, which was the date when the UN WHO declared COVID19 a pandemic. The dashed vertical orange or blue lines correspond to any significant breakpoints in the trend for GDELT or Twitter respectively, after conducting Bonferroni family-wise error correction.}
	\label{fig:sentTime}
\end{figure}

Practitioners and researchers may also seek to monitor changes in public sentiment toward different taxa. We illustrate how such analyses can be conducted with the outputs of our pipeline. Focusing on the same focal taxa of elephant, horseshoe bat, and pangolin, we compare and contrast changes in mean article or Twitter post sentiment, measured through a unidimensional value ranging from -1 (very negative) to 0 (neutral) to 1 (very positive) (Figure~\ref{fig:sentTime}). We saw that the mean sentiment toward taxa was lowest for pangolins in the news, with a mean of -0.01, and highest for long-tongued bat on Twitter, with a mean of 0.28. Of the three focal species, pangolins on the news had the lowest average sentiment, but higher sentiment on Twitter (0.14), elephants had an average sentiment of 0.12 (news) or 0.13 (Twitter), and horseshoe bats had the highest average sentiment across the board (0.18 in the news and 0.26 on Twitter).

Across all of the taxa in our pipeline, we only observed a significant breakpoint in sentiment for horseshoe bat (Figure~\ref{fig:SIsentiment}). Horseshoe bat discourse showed a change in sentiment on October 6, 2020 across both news media coverage and Twitter posts. The mean sentiment of horseshoe bat coverage remained the same (0.2 on a scale from -1 to +1) in the news media; however, Twitter horseshoe bat sentiment changed from an average of 0.2 to 0.4, becoming more positive through time.

\section{Discussion}

Our data pipeline permits practitioners and researchers to monitor public perceptions of biodiversity globally and with geographic or temporal disaggregation. This project builds on several recent advances using NLP machine learning approaches to process and analyze large, unstructured text data about biodiversity. For instance, \textcite{kulkarniAutomatedRetrievalInformation2021} created a pipeline to detect and extract news articles mentioning more than 500 CITES Appendix 1 species in the news media. \textcite{egriDetectingHotspotsHumanWildlife2022} analyzed articles from the Times of India for instances of human-wildlife conflict in West Bengal with 15 species, such as the Asian elephant (\textit{Elephas maximas}). In our project, we extended these bodies of work by simultaneously scraping data from the news media and social media, by creating a folk taxonomy to broaden the data sampled, by leveraging cutting-edge large language models to filter our data in an efficient, performant, and replicable fashion, and by scraping the Internet Archive, which permits us to mitigate issues such as the ephemerality of digital media. 

One advance of our approach is using approaches drawn from string algorithms and graph theory to generate a folk taxonomy. We showed examples of how researchers and practitioners can develop a folk taxonomy to identify unique common names for sets of species. Such a folk taxonomy can permit future monitoring efforts to capture more of the potentially relevant discourse toward biodiversity \parencite{beaudreau2011}. At the same time, using a folk taxonomy increases the volume of results at the expense of introducing some quantity of non-relevant content; we demonstrate how zero-shot LLMs can deal with this problem. 

Zero-shot approaches allow conservationists to judiciously and efficiently filter public content about biodiversity using cutting-edge machine learning models ``out of the box''. Therefore, conservation practitioners and researchers do not need to invest resources in creating labelled training and test data, a necessity for training a model from the ground up or fine-tuning an existing model using transfer learning, which may not always be feasible or advisable, especially for complex models with millions of parameters. In our case, we found that up to 62\% of the articles about bats were irrelevant, showing the importance of filtering out results that are not related to public perceptions of nature. The choropleth maps gesture toward the relative popularity of different folk taxa; we observed that there was much broader coverage and higher volume in general for prominent taxa such as elephants or more generalized entities such as bat across Twitter and news media.

Using the data generated from our pipeline, we saw that taxa differed in the distribution of topics in news media articles, and in terms of their volume and sentiment through time in the news media or on Twitter. We saw that the volume of news articles about horseshoe bats increased during the early days of the COVID19 outbreak, and that this contrasted with the other bat taxa in our pipeline, which largely did not exhibit any significant changes in volume. In contrast, while sentiment toward horseshoe bats was generally positive, our results indicated that there was a significant breakpoint in sentiment for both Twitter and news media horseshoe bat discourse in late 2020. These types of analyses can be extended in future monitoring and research efforts to evaluate the impact of public campaigns to conserve biodiversity or monitor human-nature perceptions in general \parencite{fernandez2020natural,hammond2022,millard2021a,wrightOnlineMonitoringGlobal2020,correiaDigitalDataSources2021}. Changes in the volume of discourse about species can herald problems such as the societal extinction of rare species \parencite{jaric2022}. Calculating metrics such as volume and sentiment from automated data tracking public perceptions of biodiversity offers new, standardized ways to monitor public interest in biodiversity more broadly \parencite{deoliveiracaetano2022}.

Our approach can serve as the foundation for an automated ``nature tracker'', which would permit practitioners and researchers to track public perceptions of biodiversity. Monitoring human-nature perceptions is critical to evaluating progress toward the targets of the Global Biodiversity Framework, particularly the targets focused on human-wildlife conflict and sustainable use. By scraping and processing data from the news and social media, we can provide real-time, cost-effective insight that is global in scale. Therefore, digital approaches open new avenues for assessing compliance with the Global Biodiversity Framework, which is particularly pressing given that even as late as 2024 most of these targets still show significant gaps in their evaluation mechanisms. Without effective resolution that enables tracking over substantial time periods, these monitoring deficiencies could render the targets politically irrelevant, as it would be impossible to evaluate the progress--or lack thereof--made by different countries.

In considering the future development of our data pipeline, we have identified several key areas for future exploration and enhancement. A primary aspect to address revolves around the linguistic scope of our approach, which currently centers solely on English-language data. It will be key for future work to broaden this scope to include other languages spoken in megadiverse countries, such as Spanish, Chinese, Portuguese or Bahasa Indonesia. Furthermore, it is evident that our conservation social science monitoring must adapt to dynamic shifts in platform governance and data accessibility. Recent transitions in the ownership and management of platforms like Twitter have underscored the urgency of this need. These transitions have coincided with the proliferation of misinformation regarding climate change \parencite{isdcaad23} and wildlife in the context of the COVID-19 pandemic and a marked decline in active users, particularly environmentally-focused users \parencite{MITtechrev22,chang2023environmental}, both of which pose increasing challenges to monitoring approaches using online data.

Overall, this study highlights the potential benefits of combining machine learning with the automated tracking of different data platforms to monitor public perceptions of biodiversity. We anticipate that methods such as ours or building on our approach can enhance applied conservation by creating new ways to examine human-nature perceptions at a global scale.

\section{Acknowledgments}
We express our gratitude to the institutions that supported the work presented in this paper, which was conducted while the authors were affiliated as listed in the manuscript. We note that several authors have since moved to new affiliations: Tony Chang is now with Vibrant Planet, Amrita Gupta is at the Microsoft AI for Good Lab, Diogo Verissimo is currently at the University of Oxford, and Noah Giebink is with the Spatial Informatics Group. These changes in affiliation are mentioned for the sake of accuracy regarding the authors’ current positions.

\newpage
\linespread{1}
\nolinenumbers
\printbibliography

@article{beaudreau2011,
  title = {Using Folk Taxonomies to Understand Stakeholder Perceptions for Species Conservation},
  author = {Beaudreau, Anne H. and Levin, Phillip S. and Norman, Karma C.},
  year = {2011},
  journal = {Conservation Letters},
  volume = {4},
  number = {6},
  pages = {451--463},
  issn = {1755-263X},
  urldate = {2023-05-26}
}

@article{burivalova2018,
  title = {Analyzing {{Google}} Search Data to Debunk Myths about the Public's Interest in Conservation},
  author = {Burivalova, Zuzana and Butler, Rhett A and Wilcove, David S},
  year = {2018},
  month = nov,
  journal = {Frontiers in Ecology and the Environment},
  volume = {16},
  number = {9},
  pages = {509--514},
  issn = {15409295},
  urldate = {2023-03-07}
}

@misc{CBD2022KumingMontreal,
    author = {{Convention on Biological Diversity}},
    title = {Kunming-Montreal Global Biodiversity Framework},
    year = {2022},
    url = {https://www.cbd.int/doc/decisions/cop-15/cop-15-dec-04-en.pdf},
    
}

@article{changBioScience22,
  title = {Environmental Discourse Exhibits Consistency and Variation across Spatial Scales on {{Twitter}}},
  author = {Chang, Charlotte H and Masuda, Yuta J and Armsworth, Paul R},
  year = {2022},
  journal = {BioScience},
  volume = {72},
  number = {8},
  pages = {789--797}
}

@article{chang2023environmental,
  title={{Environmental users abandoned Twitter after Musk takeover}},
  author={Chang, Charlotte H and Deshmukh, Nikhil R and Armsworth, Paul R and Masuda, Yuta J},
  journal={Trends in Ecology \& Evolution},
  year={2023},
  publisher={Elsevier}
}

@article{cooper2019,
  title = {Developing a Global Indicator for {{Aichi Target}} 1 by Merging Online Data Sources to Measure Biodiversity Awareness and Engagement},
  author = {Cooper, Matthew W. and Di Minin, Enrico and Hausmann, Anna and Qin, Siyu and Schwartz, Aaron J. and Correia, Ricardo Aleixo},
  year = {2019},
  month = feb,
  journal = {Biological Conservation},
  volume = {230},
  pages = {29--36},
  issn = {0006-3207},
  urldate = {2023-03-02}
}

@article{correiaDigitalDataSources2021,
  title = {Digital Data Sources and Methods for Conservation Culturomics},
  author = {Correia, Ricardo A. and Ladle, Richard and Jari{\'c}, Ivan and Malhado, Ana C. M. and Mittermeier, John C. and Roll, Uri and {Soriano-Redondo}, Andrea and Ver{\'i}ssimo, Diogo and Fink, Christoph and Hausmann, Anna and {Guedes-Santos}, Jhonatan and Vardi, Reut and Di Minin, Enrico},
  year = {2021},
  journal = {Conservation Biology},
  volume = {35},
  number = {2},
  pages = {398--411},
  issn = {1523-1739},
  urldate = {2022-08-12}
}

@article{deoliveiracaetano2022,
  title = {Evaluating Global Interest in Biodiversity and Conservation},
  author = {{de Oliveira Caetano}, Gabriel Henrique and {vardi}, Reut and Jari{\'c}, Ivan and Correia, Ricardo A. and Roll, Uri and Ver{\'i}ssimo, Diogo},
  year = {2022},
  month = mar,
  journal = {Conservation Biology},
  urldate = {2023-04-11}
}

@article{diaz2019,
  title = {Pervasive Human-Driven Decline of Life on {{Earth}} Points to the Need for Transformative Change},
  author = {D{\'i}az, Sandra and Settele, Josef and Brond{\'i}zio, Eduardo S. and Ngo, Hien T. and Agard, John and Arneth, Almut and Balvanera, Patricia and Brauman, Kate A. and Butchart, Stuart H. M. and Chan, Kai M. A. and Garibaldi, Lucas A. and Ichii, Kazuhito and Liu, Jianguo and Subramanian, Suneetha M. and Midgley, Guy F. and Miloslavich, Patricia and Moln{\'a}r, Zsolt and Obura, David and Pfaff, Alexander and Polasky, Stephen and Purvis, Andy and Razzaque, Jona and Reyers, Belinda and Chowdhury, Rinku Roy and Shin, Yunne-Jai and {Visseren-Hamakers}, Ingrid and Willis, Katherine J. and Zayas, Cynthia N.},
  year = {2019},
  month = dec,
  journal = {Science},
  volume = {366},
  number = {6471},
  pages = {eaax3100},
  publisher = {{American Association for the Advancement of Science}},
  urldate = {2023-02-24}
}

@inproceedings{egriDetectingHotspotsHumanWildlife2022,
  title = {Detecting {{Hotspots}} of {{Human-Wildlife Conflicts}} in {{India}} Using {{News Articles}} and {{Aerial Images}}},
  booktitle = {{{ACM SIGCAS}}/{{SIGCHI Conference}} on {{Computing}} and {{Sustainable Societies}} ({{COMPASS}})},
  author = {Egri, Gokhan and Han, Xinran and Ma, Zilin and Surapaneni, Priyanka and Chakraborty, Sunandan},
  year = {2022},
  month = jun,
  pages = {375--385},
  publisher = {{ACM}},
  address = {{Seattle WA USA}},
  urldate = {2022-09-02},
  isbn = {978-1-4503-9347-8}
}

@article{fernandez2020natural,
  title = {Natural History Films Raise Species Awareness\textemdash{{A}} Big Data Approach},
  author = {{Fern{\'a}ndez-Bellon}, Dar{\'i}o and Kane, Adam},
  year = {2020},
  journal = {Conservation Letters},
  volume = {13},
  number = {1},
  pages = {e12678},
  publisher = {{Wiley Online Library}}
}

@article{fink2020,
  title = {Online Sentiment towards Iconic Species},
  author = {Fink, Christoph and Hausmann, Anna and Di Minin, Enrico},
  year = {2020},
  month = jan,
  journal = {Biological Conservation},
  volume = {241},
  pages = {108289},
  issn = {0006-3207},
  urldate = {2023-03-02}
}

@article{hammond2022,
  title = {Examining Attention given to Threats to Elephant Conservation on Social Media},
  author = {Hammond, Niall L. and Dickman, Amy and Biggs, Duan},
  year = {2022},
  journal = {Conservation Science and Practice},
  volume = {4},
  number = {10},
  pages = {e12785},
  issn = {2578-4854},
  urldate = {2022-08-12}
}

@article{hunter2023using,
  title={Using hierarchical text classification to investigate the utility of machine learning in automating online analyses of wildlife exploitation},
  author={Hunter, Sara Bronwen and Mathews, Fiona and Weeds, Julie},
  journal={Ecological Informatics},
  volume={75},
  pages={102076},
  year={2023},
  publisher={Elsevier}
}

@inproceedings{hutto2014vader,
  title = {Vader: {{A}} Parsimonious Rule-Based Model for Sentiment Analysis of Social Media Text},
  booktitle = {Proceedings of the International {{AAAI}} Conference on Web and Social Media},
  author = {Hutto, Clayton and Gilbert, Eric},
  year = {2014},
  volume = {8},
  pages = {216--225}
}

@techreport{isdcaad23,
  type = {Report},
  title = {Deny, Deceive, Delay Vol. 2: {{Exposing}} New Trends in Climate Mis- and Disinformation at {{COP27}}},
  author = {King, Jennie},
  year = {2023},
  institution = {{Climate Action Against Disinformation, Institute for Strategic Dialogue}}
}

@article{jaric2020,
  title = {Societal Attention toward Extinction Threats: A Comparison between Climate Change and Biological Invasions},
  author = {Jari{\'c}, Ivan and Bellard, C{\'e}line and Courchamp, Franck and Kalinkat, Gregor and Meinard, Yves and Roberts, David L. and Correia, Ricardo A.},
  year = {2020},
  month = jul,
  journal = {Scientific Reports},
  volume = {10},
  number = {1},
  pages = {11085},
  publisher = {{Nature Publishing Group}},
  issn = {2045-2322},
  urldate = {2023-03-02},
  copyright = {2020 The Author(s)}
}

@article{jaric2022,
  title = {Societal Extinction of Species},
  author = {Jari{\'c}, Ivan and Roll, Uri and Bonaiuto, Marino and Brook, Barry W. and Courchamp, Franck and Firth, Josh A. and Gaston, Kevin J. and Heger, Tina and Jeschke, Jonathan M. and Ladle, Richard J. and Meinard, Yves and Roberts, David L. and Sherren, Kate and Soga, Masashi and {Soriano-Redondo}, Andrea and Ver{\'i}ssimo, Diogo and Correia, Ricardo A.},
  year = {2022},
  month = may,
  journal = {Trends in Ecology \& Evolution},
  volume = {37},
  number = {5},
  pages = {411--419},
  issn = {0169-5347},
  urldate = {2023-02-10}
}

@article{keh2023NewsPanda,
  title = {{{NewsPanda}}: {{Media Monitoring}} for {{Timely Conservation Action}}},
  shorttitle = {{{NewsPanda}}},
  author = {Keh, Sedrick Scott and Shi, Zheyuan Ryan and Patterson, David J. and Bhagabati, Nirmal and Dewan, Karun and Gopala, Areendran and Izquierdo, Pablo and Mallick, Debojyoti and Sharma, Ambika and Shrestha, Pooja and Fang, Fei},
  date = {2023-06-26},
  journaltitle = {Proceedings of the AAAI Conference on Artificial Intelligence},
  shortjournal = {AAAI},
  volume = {37},
  number = {13},
  pages = {15528--15536},
  issn = {2374-3468, 2159-5399},
  doi = {10.1609/aaai.v37i13.26841},
  url = {https://ojs.aaai.org/index.php/AAAI/article/view/26841},
  urldate = {2023-09-21},
  langid = {english},
}

@article{killick2014,
  title = {{{changepoint}}: {{An R}} Package for Changepoint Analysis},
  author = {Killick, Rebecca and Eckley, Idris A.},
  year = {2014},
  journal = {Journal of Statistical Software},
  volume = {58},
  number = {3},
  pages = {1--19}
}

@article{king2017news,
  title = {How the News Media Activate Public Expression and Influence National Agendas},
  author = {King, Gary and Schneer, Benjamin and White, Ariel},
  year = {2017},
  journal = {Science (New York, N.Y.)},
  volume = {358},
  number = {6364},
  pages = {776--780},
  publisher = {{American Association for the Advancement of Science}}
}

@article{kulkarniAutomatedRetrievalInformation2021,
  title = {Automated Retrieval of Information on Threatened Species from Online Sources Using Machine Learning},
  author = {Kulkarni, Ritwik and Di Minin, Enrico},
  year = {2021},
  journal = {Methods in Ecology and Evolution},
  volume = {12},
  number = {7},
  pages = {1226--1239},
  issn = {2041-210X},
  urldate = {2021-12-07}
}

@article{ladle2016,
  title = {Conservation Culturomics},
  author = {Ladle, Richard J and Correia, Ricardo A and Do, Yuno and Joo, Gea-Jae and Malhado, Ana CM and Proulx, Rapha{\"e}l and Roberge, Jean-Michel and Jepson, Paul},
  year = {2016},
  journal = {Frontiers in Ecology and the Environment},
  volume = {14},
  number = {5},
  pages = {269--275},
  issn = {1540-9309},
  urldate = {2023-02-14}
}

@article{ladle2019culturomics,
  title={A culturomics approach to quantifying the salience of species on the global internet},
  author={Ladle, Richard J and Jepson, Paul and Correia, Ricardo A and Malhado, Ana CM},
  journal={People Nat},
  volume={1},
  number={4},
  pages={524--532},
  year={2019}
}

@misc{lewis2019,
  title = {{{BART}}: {{Denoising Sequence-to-Sequence Pre-training}} for {{Natural Language Generation}}, {{Translation}}, and {{Comprehension}}},
  author = {Lewis, Mike and Liu, Yinhan and Goyal, Naman and Ghazvininejad, Marjan and Mohamed, Abdelrahman and Levy, Omer and Stoyanov, Ves and Zettlemoyer, Luke},
  year = {2019},
  month = oct,
  number = {arXiv:1910.13461},
  eprint = {1910.13461},
  primaryclass = {cs, stat},
  publisher = {{arXiv}},
  urldate = {2023-04-20},
  archiveprefix = {arxiv}
}

@article{millard2021a,
  ids = {millard2021},
  title = {The Species Awareness Index as a Conservation Culturomics Metric for Public Biodiversity Awareness},
  author = {Millard, Joseph W. and Gregory, Richard D. and Jones, Kate E. and Freeman, Robin},
  year = {2021},
  journal = {Conservation Biology},
  volume = {35},
  number = {2},
  pages = {472--482},
  issn = {1523-1739},
  urldate = {2022-11-19}
}

@techreport{MITtechrev22,
  type = {Report},
  title = {Twitter May Have Lost More than a Million Users since {{Elon Musk}} Took Over},
  author = {{Stokel-Walker}, Chris},
  year = {2022},
  institution = {{MIT Technology Review}}
}

@article{papworth2015,
  title = {{Quantifying the role of online news in linking conservation research to Facebook and Twitter}},
  author = {Papworth, S.k. and Nghiem, T.p.l. and Chimalakonda, D. and Posa, M.r.c. and Wijedasa, L.s. and Bickford, D. and Carrasco, L.r.},
  year = {2015},
  journal = {Conservation Biology},
  volume = {29},
  number = {3},
  pages = {825--833},
  issn = {1523-1739},
  urldate = {2023-03-02}
}

@article{petrovan2021,
  title = {Post {{COVID-19}}: A Solution Scan of Options for Preventing Future Zoonotic Epidemics},
  author = {Petrovan, Silviu O. and Aldridge, David C. and Bartlett, Harriet and Bladon, Andrew J. and Booth, Hollie and Broad, Steven and Broom, Donald M. and Burgess, Neil D. and Cleaveland, Sarah and Cunningham, Andrew A. and Ferri, Maurizio and Hinsley, Amy and Hua, Fangyuan and Hughes, Alice C. and Jones, Kate and Kelly, Moira and Mayes, George and Radakovic, Milorad and Ugwu, Chinedu A. and Uddin, Nasir and Ver{\'i}ssimo, Diogo and Walzer, Christian and White, Thomas B. and Wood, James L. and Sutherland, William J.},
  year = {2021},
  journal = {Biological Reviews},
  volume = {96},
  number = {6},
  pages = {2694--2715},
  issn = {1469-185X},
  urldate = {2023-03-07}
}

@article{roberge2014,
  title = {Using Data from Online Social Networks in Conservation Science: Which Species Engage People the Most on {{Twitter}}?},
  author = {Roberge, Jean-Michel},
  year = {2014},
  month = mar,
  journal = {Biodiversity and Conservation},
  volume = {23},
  number = {3},
  pages = {715--726},
  issn = {1572-9710},
  urldate = {2023-03-02}
}

@article{roll2018using,
  title={Using machine learning to disentangle homonyms in large text corpora},
  author={Roll, Uri and Correia, Ricardo A and Berger-Tal, Oded},
  journal={Conservation Biology},
  volume={32},
  number={3},
  pages={716--724},
  year={2018},
  publisher={Wiley Online Library}
}

@article{thaler2017lions,
  title={Lions, whales, and the web: transforming moment inertia into conservation action},
  author={Thaler, Andrew D and Rose, Naomi A and Cosentino, A Mel and Wright, Andrew J},
  journal={Frontiers in Marine Science},
  volume={4},
  pages={292},
  year={2017},
  publisher={Frontiers Media SA}
}

@article{vardi2021,
  title = {Combining Culturomic Sources to Uncover Trends in Popularity and Seasonal Interest in Plants},
  author = {Vardi, Reut and Mittermeier, John C. and Roll, Uri},
  year = {2021},
  journal = {Conservation Biology},
  volume = {35},
  number = {2},
  pages = {460--471},
  issn = {1523-1739},
  urldate = {2023-04-11}
}

@misc{verissimo2021,
  type = {{{SSRN Scholarly Paper}}},
  title = {Trends in {{Digital Marketing}} for {{Biodiversity Conservation}}},
  author = {Ver{\'i}ssimo, Diogo},
  year = {2021},
  month = may,
  number = {3878918},
  address = {{Rochester, NY}},
  urldate = {2023-02-13}
}

@article{vijay2021,
  title = {Using Internet Search Data to Understand Information Seeking Behavior for Health and Conservation Topics during the {{COVID-19}} Pandemic},
  author = {Vijay, Varsha and Field, Christopher R. and Gollnow, Florian and Jones, Kelly K.},
  year = {2021},
  month = may,
  journal = {Biological Conservation},
  volume = {257},
  pages = {109078},
  issn = {0006-3207},
  urldate = {2023-03-07}
}

@misc{wrightOnlineMonitoringGlobal2020,
  title = {Online {{Monitoring}} of {{Global Attitudes Towards Wildlife}}},
  author = {Wright, Joss and Lennox, Robert and Ver{\'i}ssimo, Diogo},
  year = {2020},
  month = jul,
  number = {arXiv:2007.11506},
  eprint = {2007.11506},
  primaryclass = {cs},
  publisher = {{arXiv}},
  urldate = {2022-09-02},
  archiveprefix = {arxiv}
}

@article{zhou2020pneumonia,
  title = {A Pneumonia Outbreak Associated with a New Coronavirus of Probable Bat Origin},
  author = {Zhou, Peng and Yang, Xing-Lou and Wang, Xian-Guang and Hu, Ben and Zhang, Lei and Zhang, Wei and Si, Hao-Rui and Zhu, Yan and Li, Bei and Huang, Chao-Lin and others},
  year = {2020},
  journal = {nature},
  volume = {579},
  number = {7798},
  pages = {270--273},
  publisher = {{Nature Publishing Group}}
}

\newpage

\appendix
\setcounter{table}{0}
\renewcommand{\thetable}{A\arabic{table}}

\section{Supplementary Information}
\subsection{Folk Taxonomy Graph Construction}
\label{subsec:appendix-folktax-graph}
To construct the graph, a node is created for every:
\begin{itemize}
    \item scientific name for a species (e.g. \emph{Rhinolophus affinis})
    \item full common name for a species (e.g. ``Intermediate horseshoe bat'')
    \item substring shared between multiple common names (e.g. ``Horseshoe bat'' is part of ``Intermediate horseshoe bat'' and ``Greater horseshoe bat'') or substrings (e.g. ``Bat'' is part of ``Horseshoe bat'' and ``Fruit bat'')
\end{itemize}
Edges exist between the nodes representing the scientific name and the full common names for each species, as well as between common names and their substrings.

Table~\ref{tab:folktax} shows the folk taxonomy entries that we designed using graph theory applied to species' common names.

\begin{table}[!htbp]
	\caption{Table mapping folk taxonomy entities to scientific taxa.}
	\label{tab:folktax}
	\begin{tabular}{lp{0.2\linewidth}lp{0.25\linewidth}p{0.25\linewidth}}
		\toprule
		Order & Scientific taxon & Folk taxon & Positive keywords & Negative keywords\\
		\midrule
		Chiroptera & Order Chiroptera & Bat & bat \\
		Chiroptera & Family Pteropodidae & Flying fox & flying fox, pale xantharpy,  acerodon, monkey-faced bat, greater nectar bat, fruit bat, north moluccan blossum-bat, rousette, golden bat of rodrigues, codot horsfield, woerman's bat, blossom bat & tube-nosed fruit bat\\
		Chiroptera & Family Hipposideridae,
		Family Rhinonlophidae & Horseshoe bat &
diadem leafnosed-bat, roundleaf bat, horseshoe-bat, horseshoe bat, great woolly horsehoe bat, trident bat, leaf-nosed bat, flower-faced bat \\
		Chiroptera & Genus Glossophaga,
		Genus Craseonycteris,
		Genus Leptonycteris & Long tongued bat  & hog-nosed bat, long-nosed bat, bumblebee bat, long-tongued bat\\
		Chiroptera & Genus Myotis,
		Genus Perimyotis & Myotis & pond bat, bocage's banana bat, van hasselts bat, social bat, bechstein's bat, hodgson's bat, lesser large-tooth bat, mouse-eared bat, ridley's bat, water bat, ikonnikov's bat, whiskered bat, welwitch's bat, three-coloured bat, daubenton's bat, indiana bat, fish-eating bat, grey bat, siliguri bat, geoffroy's bat, large-footed bat, myotis, horsfield's bat, rickett's big-footed bat, little brown bat, brandt's bat, siberian bat, morris's bat, natterer's bat, intermediate bat, hairy-faced bat, Descaleras bat\\
		Chiroptera & Subfamily Vespertilioninae & Pipistrelle &  pipistrelle, pipistrelle bat, cape bat, anchieta's bat, rohus bat, thai golden-throated bat, siam goldnecklet, dormer's bat, ruppell's bat, rusty bat, little indian bat, white-winged bat, sind bat, aloe bat, hottentot bat, indian pygmy bat, banana bat & bocages banana bat\\
		Chiroptera & Family Emballonuridae,	Family Phyllostomidae, Family Megadermatidae & Vampire bat & ghost bat, vampire bat, false vampire, heart-nosed bat\\
		Proboscidea & Family Elephantidae & Elephant & elephant\\ 
		Pholidota & Family Manidae & Pangolin & pangolin\\
		Primates & Genus Gorilla & Gorilla & gorilla \\
		\bottomrule
	\end{tabular}
\end{table}

Table~\ref{tab:relevance} displays the counts of news media articles for different taxa classified as relevant verses irrelevant to biodiversity and categorized as original or syndicated.

\begin{table}[!htbp]
	\caption{The counts across taxa for original versus duplicate articles identified in our pipeline and relevant articles determined by zero-shot learning with a transformer model.}
	\label{tab:relevance}
	\begin{tabular}{llrr}
		\toprule 
		& Predicted relevance: & True & False \\
		Taxon & Original article &  &  \\
		\midrule
		\multirow[t]{2}{*}{Elephant} & True & 86425 & 70548 \\
		& False & 69335 & 54403 \\
		\cline{1-4}
		\multirow[t]{2}{*}{Gorilla} & True & 21399 & 21206 \\
		& False & 11715 & 9476 \\
		\cline{1-4}
		\multirow[t]{2}{*}{Pangolin} & True & 6286 & 1495 \\
		& False & 5976 & 1055 \\
		\cline{1-4}
		\multirow[t]{2}{*}{Bat} & True & 77802 & 129058 \\
		& False & 45452 & 68122 \\
		\cline{1-4}
		\multirow[t]{2}{*}{Flying fox} & True & 1969 & 601 \\
		& False & 1311 & 447 \\
		\cline{1-4}
		\multirow[t]{2}{*}{Myotis} & True & 1103 & 103 \\
		& False & 978 & 21 \\
		\cline{1-4}
		\multirow[t]{2}{*}{Horseshoe bat} & True & 791 & 171 \\
		& False & 745 & 257 \\
		\cline{1-4}
		\multirow[t]{2}{*}{Pipistrelle} & True & 496 & 63 \\
		& False & 489 & 45 \\
		\cline{1-4}
		\multirow[t]{2}{*}{Vampire bat} & True & 341 & 195 \\
		& False & 357 & 176 \\
		\cline{1-4}
		\multirow[t]{2}{*}{Long-tongued bat} & True & 150 & 24 \\
		& False & 79 & 9 \\
		\bottomrule
	\end{tabular}
\end{table}

\newpage

\begin{figure}
	\centering
	\includegraphics[height=0.9\textheight]{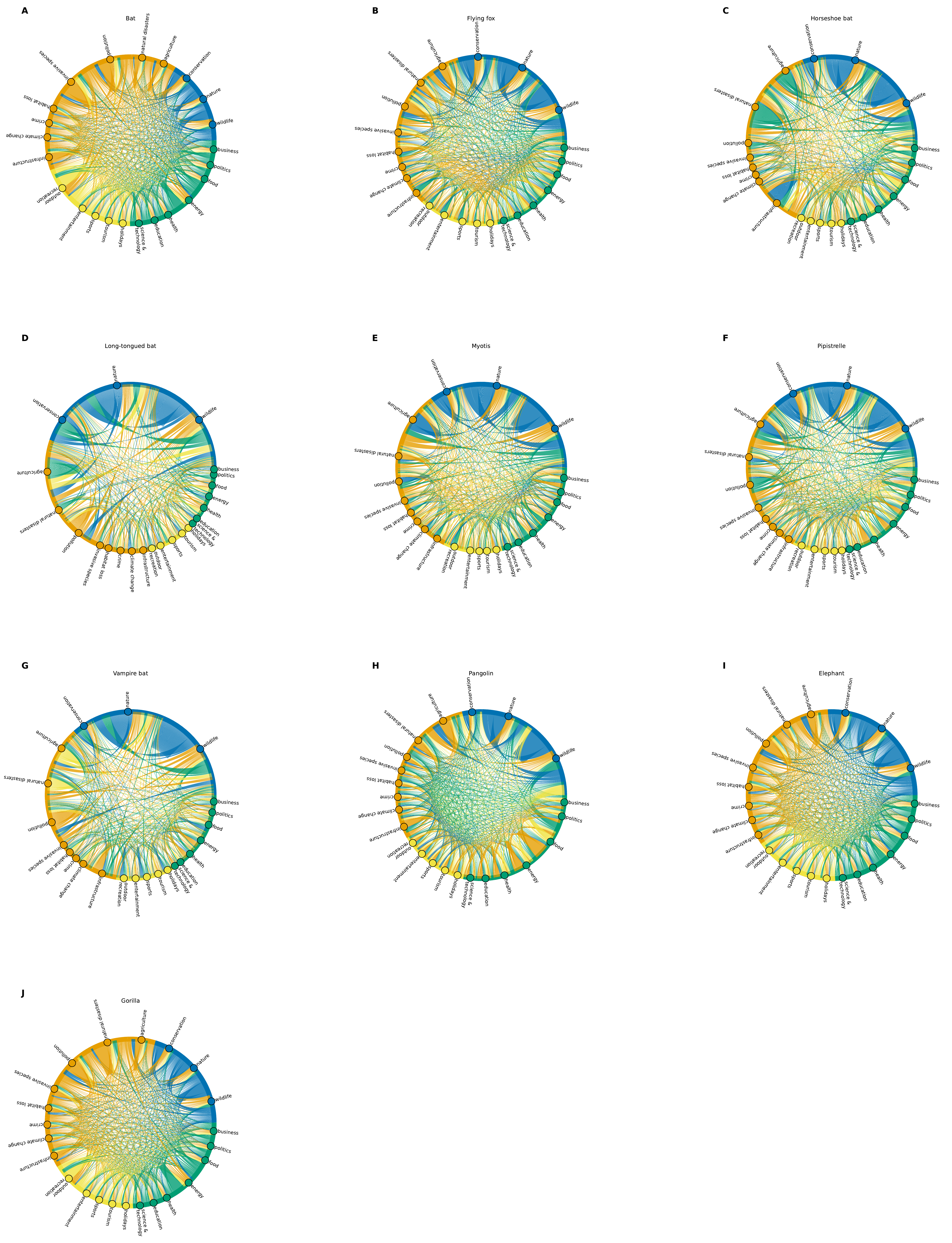}
	\caption{Chord diagrams depicting the co-occurrence of relevant topics for the different taxa.}
	\label{fig:relevantTopics}
\end{figure}

Figure~\ref{fig:relevantTopics} shows the distribution of topics associated with relevant articles for different taxa.

\begin{figure}
	\centering
	\includegraphics[width=0.9\textwidth]{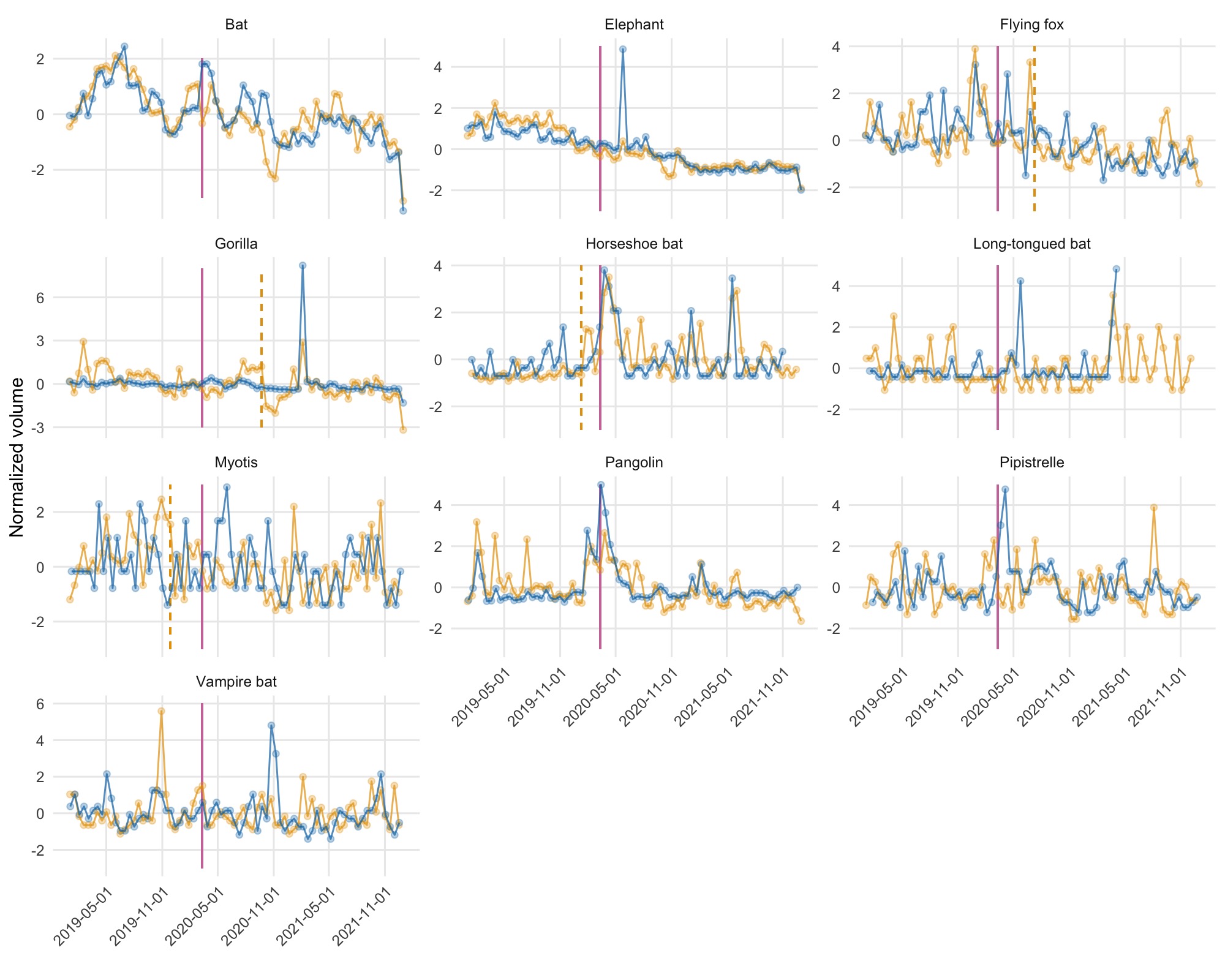}
	\caption{Changes in volume through time. The solid vertical magenta line denotes March 11, 2020, which was the date when the UN WHO declared COVID19 a pandemic. The dashed vertical orange or blue lines correspond to any significant breakpoints in the trend for GDELT or Twitter respectively, after conducting Bonferroni family-wise error correction.}
	\label{fig:SIvolume}
\end{figure}

Figure~\ref{fig:SIvolume} displays changes in volume for all of the focal taxa analyzed in the pipeline.

\begin{figure}
	\centering
	\includegraphics[width=0.9\textwidth]{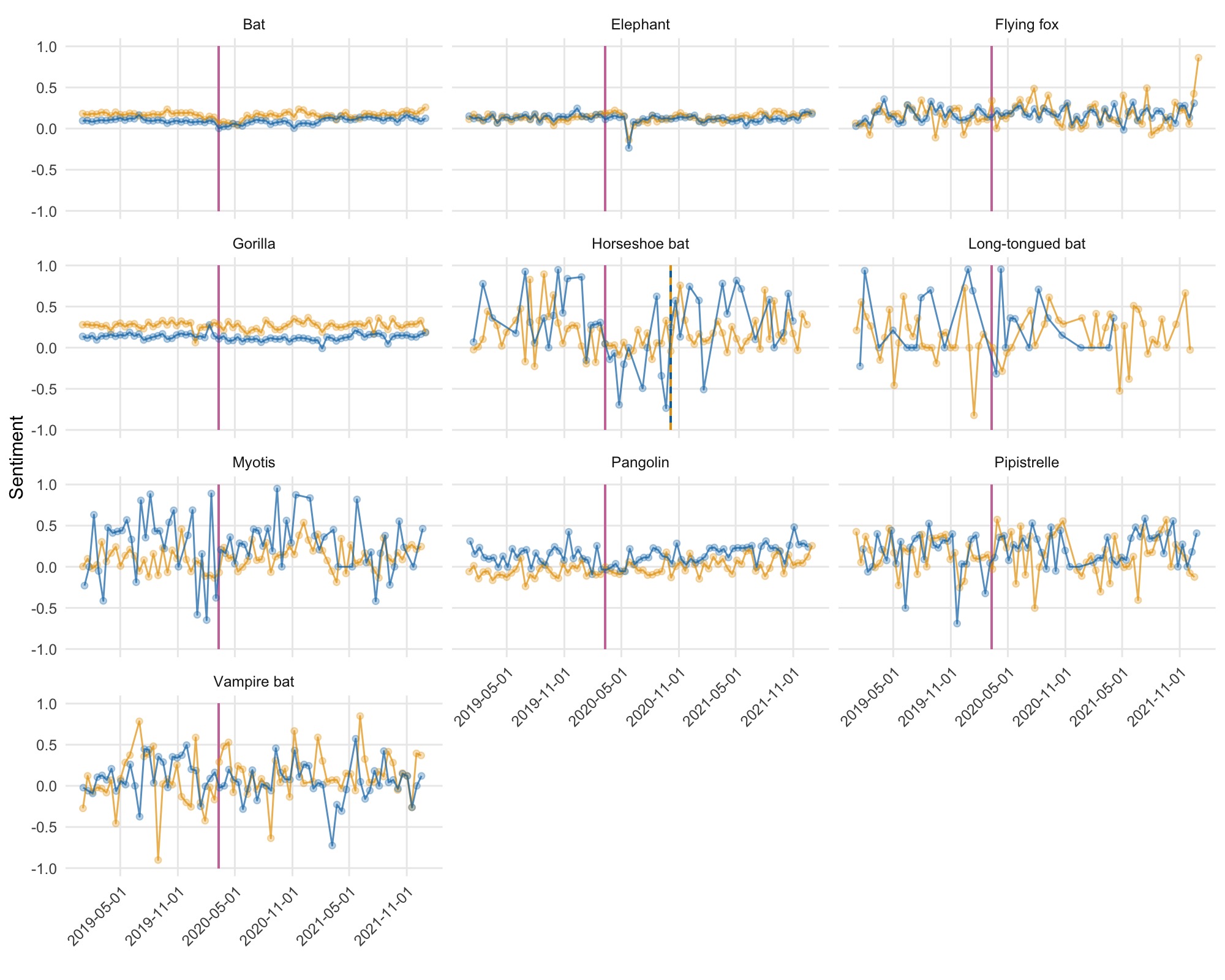}
	\caption{Changes in sentiment through time. The solid vertical magenta line denotes March 11, 2020, which was the date when the UN WHO declared COVID19 a pandemic. The dashed vertical orange or blue lines correspond to any significant breakpoints in the trend for GDELT or Twitter respectively, after conducting Bonferroni family-wise error correction.}
	\label{fig:SIsentiment}
\end{figure}

Figure~\ref{fig:SIsentiment} displays changes in mean article sentiment for all of the focal taxa analyzed in the pipeline.

\end{document}